\documentclass[journal,10pt,twocolumn]{IEEEtran}

\usepackage{amsmath,amsfonts,amssymb}
\usepackage{graphicx}
\usepackage{cite}
\usepackage{bm}
\usepackage{url}
\usepackage{acronym}
\usepackage{booktabs}
\usepackage[caption=false,font=footnotesize]{subfig}
\usepackage{algorithm}
\usepackage{algorithmicx}
\usepackage{algpseudocode}
\usepackage{enumerate}

\newacro{AWGN}{additive white Gaussian noise}
\newacro{ALU}{arithmetic logic unit}
\newacro{BCQ}{binary-coding quantization}
\newacro{CPU}{central processing unit}
\newacro{DCT}{discrete cosine transform}

\newacro{FC}{fusion center}
\newacro{FJLT}{fast Johnson-Lindenstrauss transform}
\newacro{FWHT}{fast Walsh-Hadamard transform}
\newacro{FIR}{finite impulse response}
\newacro{FPGA}{field-programmable gate array}

\newacro{GPU}{graphics processing unit}

\newacro{i.i.d}{independent and identically distributed}
\newacro{JIT}{just-in-time}
\newacro{KV}{key-value}
\newacro{LLM}{large language model}
\newacro{LUT}{look-up table}
\newacro{MSE}{mean-squared error}
\newacro{MAC}{multiply-accumulate}

\newacro{PU}{primary user}
\newacro{POPCNT}{population count}
\newacro{PRIDe}{{Pietra-Ricci} index detector}
\newacro{PID}{power iteration detector}
\newacro{PQ}{product quantization}
\newacro{PoT}{power-of-two}
\newacro{QJLT}{quantized Johnson-Lindenstrauss transform} 
\newacro{ROC}{receiver operating characteristic}
\newacro{RAG}{retrieval-augmented generation}
\newacro{SU}{secondary user}
\newacro{SCM}{sample covariance matrix}
\newacro{SPID}{sketched power iteration detector}
\newacro{SNR}{signal-to-noise ratio}
\newacro{SRAM}{static random-access memory}
\newacro{SGD}{stochastic gradient descent}

\newacro{WHT}{Walsh-Hadamard transform}

\begin{document}

\title{Fast-TurboQuant A Multiplier-Free Online Vector Quantization Approach}

\author{

    \IEEEauthorblockN{
    Pedro Pereira,
    Felipe de Figueiredo,
    Rausley de Souza
    }
    \thanks{
    
    This work was funded in part by CNPq (Grant Nos. 302085/2025-4 and 306199/2025-4), by Minas Gerais Research Foundation (FAPEMIG) (Grant Nos. PPE-00124-23, APQ-04523-23, APQ-05305-23, RED-00194-23, and APQ-03162-24), by the Brasil 6G project (01245.020548/2021-07), supported by RNP and MCTI, and by the projects XGM-AFCCT-2024-2-5-1 and XGM-AFCCT-2024-9-1-1 supported by xGMobile – EMBRAPII-Inatel Competence Center on 5G and 6G Networks, with financial resources from the PPI IoT/Manufatura 4.0 from MCTI grant number 052/2023, signed with EMBRAPII, and by FINEP (nº 1060/2 contract 01.25.0883.00).

    The authors are with the National Institute of Telecommunications (Inatel), Santa Rita do Sapucaí, MG  37536-001, Brazil (pedro.marcio@inatel.br; felipe.figueiredo@inatel.br; rausley@inatel.br).

    } 
}

\maketitle

\begin{abstract}
As large language models scale, memory bandwidth for key-value caches and retrieval-augmented generation systems becomes a critical bottleneck. While 1-bit quantization addresses this constraint, recent TurboQuant relies on dense random rotation matrices to condition the vector distribution before quantization. This projection demands millions of floating-point multiplications per embedding, making it difficult to deploy on constrained edge silicon. We introduce Fast-TurboQuant, a multiplier-free projection architecture that replaces the dense matrix with a structured fast Johnson-Lindenstrauss transform. By applying a Rademacher phase inversion followed by a fast Walsh-Hadamard transform (FWHT), the method leverages sub-Gaussian concentration to satisfy the prerequisites of scalar Lloyd-Max quantization without Gaussian projections. This substitution reduces the arithmetic complexity to only additions, eliminating hardware multipliers. Evaluation on DBpedia OpenAI-3 Large embeddings demonstrates a 19.7 times algorithmic speedup under sequential execution. Furthermore, the dimension expansion due to the FWHT zero-padding reduces the mean squared error and improves Recall@10.
\end{abstract}


\section{Introduction}

The deployment of \acp{LLM} and \ac{RAG} systems is constrained by memory bandwidth and capacity. In generative transformer architectures, the \ac{KV} cache grows linearly with sequence length, creating a bottleneck for long-context inference. To alleviate this, vector quantization techniques, such as \ac{PQ} and 1-bit quantization, have become essential for compressing high-dimensional embeddings while preserving their spatial geometry for fast inner-product retrieval~\cite{zhang2025pqcache}.

Recently, TurboQuant established a near-optimal distortion rate for online vector quantization~\cite{zandieh2025turboquant}. The authors found that while scalar quantizers minimize \ac{MSE}, they introduce bias in inner product estimates. To resolve this, TurboQuant employs a two-stage pipeline scalar quantization followed by a 1-bit \ac{QJLT} applied to the residual error. To ensure the scalar quantizer functions optimally across arbitrary data manifolds, TurboQuant applies a Gaussian random rotation matrix to the input vector, thereby forcing the transformed coordinates to follow a Beta distribution.

However, the optimality of TurboQuant shows a hardware inefficiency. The random rotation requires a quadratic number of projection operations relative to the embedding dimension. For high-dimensional embeddings, this preprocessing step demands millions of floating-point multiplications per token. In streaming inference applications and edge \ac{CPU} deployments, where the \ac{ALU} multiplier pipelines are constrained, this overhead neutralizes the latency benefits of the subsequent 1-bit quantization.

In this letter, we resolve this computational bottleneck by introducing Fast-TurboQuant. We demonstrate that the rotation matrix can be replaced by a structured \ac{FJLT} without sacrificing the distortion bounds. By applying a diagonal Rademacher matrix for random phase inversion, followed by the \ac{FWHT}, our architecture computes the necessary uniform rotation matrix in a matrix-free manner.

Because the phase inversion requires only sign-bit alternating, and the \ac{FWHT} evaluates through an in-place butterfly network, Fast-TurboQuant reduces the projection complexity to a log-linear number of additions. Our method eliminates hardware multipliers from the projection path. Empirical evaluations on \ac{LLM} embeddings confirm that Fast-TurboQuant achieves similar inner product distortion limits as the TurboQuant, making it desirable for edge silicon implementation.

\section{Related Work}
\label{sec:related_work}

\subsection{The Hardware Multiplier Bottleneck}
Deploying high-dimensional models on edge devices introduces memory bandwidth and \ac{ALU} constraints. Previous research seeks to eliminate floating-point multiplications from the inference data path. GoQuant approaches the issue of offline static weights by applying an orthogonal residual projection to repair angular gaps in low-bit logarithmic lattices, relying on shift-and-add hardware logic~\cite{xiang2026orpquant}. ShiftAddLLM replaces weight-activation operations with bitwise shifts and look-up tables~\cite{you2024shiftaddllm}. While these methods compress static weights for offline calibration, Fast-TurboQuant isolates the problem of online streaming vector quantization for dynamic \ac{KV} caches.

\subsection{Orthogonal Rotations and Outlier Suppression}
Heavy outliers within neural network embeddings degrade scalar quantization. TurboQuant introduced dense Gaussian matrices to rotate the data manifold, forcing coordinates into normal distributions to meet the prerequisites of optimal scalar codebooks~\cite{zandieh2025turboquant}. However, the projection needs $\mathcal{O}(d^2)$ floating-point operations.

Recent literature mitigates the dense computation via alternative techniques. SpinQuant treats rotation as a machine learning optimization challenge, using Cayley stochastic gradient descent on the Stiefel manifold to optimize rotation matrices~\cite{liu2025spinquant}. QuaRot fuses randomized Hadamard matrices into the model parameters offline~\cite{ashkboos2024quarot}. The technique maintains computational invariance while eliminating outliers inside INT4 Tensor Cores during inference. PolarQuant applies block-wise $L_2$ normalization and Hadamard rotations to chunked weight tensors~\cite{vicentino2026polarquant}. KVLinC applies Hadamard rotations to value vectors and trains linear adapters to fix generation errors caused by 2-bit quantization~\cite{saxena2025kvlinc}. These solutions target offline weight calibration or cloud-based \ac{GPU} execution.

\subsection{Transform-Domain Quantization}
The \ac{FWHT} provides a structured way to achieve sub-Gaussian concentration. Fang et al. offer mathematical proofs establishing that a randomized Hadamard transform yields the asymptotic normality required for quantization~\cite{feng2026provable}. The theoretical bounds rely on a dithering mechanism to guarantee unbiased limits.

Other implementations adapt the transform for specific domains. ITQ3\_S utilizes a deterministic 256-point transform for 3-bit static weight compression, fusing the operation into \ac{GPU} shared memory~\cite{yoon2026itq3_s}. 
RaBitQ applies shifted grids of unsigned integers to build uniform codebooks~\cite{gao2024rabitq}. The model establishes an optimal space-distortion bound where the bit-width scales as $\log\log(1/\delta)$. Symmetric evaluations show RaBitQ matches or outperforms the baseline TurboQuant in quantization speed and inner-product recall on \ac{GPU} hardware~\cite{gao2026revisiting}. Furthermore, the fast variant of RaBitQ integrates the \ac{FWHT} to bypass dense rotations. While RaBitQ provides optimal bounds for cloud infrastructure, Fast-TurboQuant shifts the design to multiplier-free \ac{ALU} paths for edge silicon.

\subsection{Fast-TurboQuant Architectural Novelty}
Fast-TurboQuant eliminates \ac{ALU} multipliers by replacing the dense rotation with a \ac{FJLT}. The Rademacher phase inversion executes via sign-bit toggles.

Furthermore, the proposed method explores the power-of-two constraint of the butterfly network. Expanding the native dimension from $d=1536$ to $N=2048$ expands the sketch dimension. The padding increases the bit capacity for the subsequent 1-bit codebook, yielding lower \ac{MSE} and better retrieval accuracy without the usage of dense memory storage. By absorbing the scalar normalization into the offline codebook design, the pipeline avoids runtime floating-point scaling.

\section{Multiplier-Free Projection Architecture}

Let $\mathbf{x} \in \mathbb{R}^d$ denote a high-dimensional input vector, such as a token embedding from a \ac{KV} cache. The TurboQuant architecture assumes the coordinates of the projected vector approximate a normal distribution $\mathcal{N}(0, 1/d)$. That statistical condition allows the system to apply a 1-dimensional Lloyd-Max scalar quantizer to each coordinate without cross-dependency. To induce the necessary distribution, the baseline formulation projects $\mathbf{x}$ using a dense Gaussian rotation matrix $\mathbf{\Pi} \in \mathbb{R}^{d \times d}$~\cite{zandieh2025turboquant}.

We replace the rotation $\mathbf{\Pi}$ with a \ac{FJLT}~\cite{ailon2009fast}. Because the \ac{FWHT} requires a power-of-two dimension, we zero-pad the input vector to length $N = 2^{\lceil \log_2 d \rceil}$, forming the expanded vector $\tilde{\mathbf{x}} \in \mathbb{R}^N$. The modified projection computes the intermediate vector $\mathbf{y} \in \mathbb{R}^N$ as
\(
\mathbf{y} = \frac{1}{\sqrt{N}} \mathbf{H} \mathbf{D} \tilde{\mathbf{x}},
\)
where $\mathbf{D} \in \{-1, 1\}^{N \times N}$ is a diagonal matrix of \ac{i.i.d} Rademacher variables, and $\mathbf{H} \in \{-1, 1\}^{N \times N}$ is the Walsh-Hadamard matrix.

While the TurboQuant architecture relies on the rotational invariance of dense matrices to achieve its coordinate distribution, the structured transform $\mathbf{H}\mathbf{D}$ relies on sub-Gaussian concentration. The Rademacher diagonal $\mathbf{D}$ acts as a random phase inverter. For any input vector $\tilde{\mathbf{x}}$, the vector $\mathbf{z} = \mathbf{D}\tilde{\mathbf{x}}$ has zero mean and an isotropic covariance matrix. From a hardware perspective, multiplying by $\mathbf{D}$ requires zero \ac{ALU} multipliers; the operation evaluates through simple sign-bit toggles.

The Walsh-Hadamard matrix $\mathbf{H}$ acts as a mixing operator. Each coordinate $y_i$ of the output vector $\mathbf{y}$ is a linear combination of all $N$ elements of $\mathbf{z}$, given by
\(
y_i = \frac{1}{\sqrt{N}} \sum_{j=1}^{N} H_{i,j} z_j.
\)
Because the components $z_j$ are zero-mean bounded variables, Hoeffding's inequality dictates that the sum exhibits sub-Gaussian tails~\cite{hoeffding1963probability}. Formal mathematical proofs establishing that the randomized Hadamard projection achieves the necessary asymptotic normality to match genuine random rotations are provided by Feng et al.~\cite{feng2026provable}. Furthermore, executing the multiplication via the \ac{FWHT} butterfly network reduces the operation to pure additions and subtractions, completing the multiplier-free data path.

\section{Complexity and Hardware Analysis}

This section quantifies the computational and memory footprint reductions achieved by Fast-TurboQuant.

\subsection{Arithmetic Complexity}
The TurboQuant architecture projects the input vector using a random rotation matrix. The matrix-vector multiplication computes $y_i = \sum_{j=1}^{d} \Pi_{i,j} x_j$, requiring $d^2$ floating-point multiplications and $d(d-1)$ floating-point additions per vector.

Fast-TurboQuant computes the projection via $\mathbf{y} = \frac{1}{\sqrt{N}} \mathbf{H} \mathbf{D} \tilde{\mathbf{x}}$. We assume the native dimension $d$ is zero-padded to the nearest power of two, denoted as $N = 2^{\lceil \log_2 d \rceil}$, forming the expanded vector $\tilde{\mathbf{x}} \in \mathbb{R}^N$. The arithmetic operations break down as follows:

\begin{enumerate}[i)]
    \item Applying the Rademacher diagonal matrix requires $N$ sign-bit inversions. The inversion executes via XOR gates on the sign bit of the floating-point representation, consuming zero \ac{ALU} cycles.
    \item The \ac{FWHT} evaluates the matrix-vector product using a divide-and-conquer butterfly network. The routing requires $N \log_2 N$ additions and subtractions, bypassing multiplications.
    \item The scalar normalization is incorporated into the subsequent 1-bit quantization thresholds or scalar Lloyd-Max codebooks during offline design, eliminating the need for runtime multiplication.
\end{enumerate}

Fast-TurboQuant reduces the arithmetic complexity of the projection from $\mathcal{O}(d^2)$ multipliers to zero multipliers and $\mathcal{O}(N \log_2 N)$ adders.

\subsection{Memory Bandwidth and Storage}
TurboQuant requires reading and storing the matrix $\mathbf{\Pi}$. For a dimension $d$, the footprint consumes $d^2$ parameters. At 16-bit precision, an embedding of dimension $d=3072$ requires $18.8$ MB of \ac{SRAM} to store the projection weights. The data volume introduces a memory bandwidth bottleneck during streaming inference.

Fast-TurboQuant is a matrix-free architecture. The Walsh-Hadamard matrix $\mathbf{H}$ operates through the fixed routing of the butterfly network and requires no memory storage. The Rademacher diagonal $\mathbf{D}$ is stored as a 1-bit vector of length $N$. For $N=4096$, the projection storage requirement drops from megabytes to 512 bytes.

\section{Evaluation}
To evaluate the proposed architecture, we benchmarked inner product estimation using the DBpedia entities dataset with OpenAI-3-Large text embeddings~\cite{dbpediaopenai3}.

\subsection{Data Preparation}We sampled a database matrix $\mathbf{X} \in \mathbb{R}^{5000 \times d}$ and a query matrix $\mathbf{Q} \in \mathbb{R}^{500 \times d}$, where the embedding dimension is $d=1536$. All vectors were $L_2$-normalized. To satisfy the power-of-two constraint of the \ac{FWHT}, the input vectors for the proposed method were zero-padded to $N=2048$. The baseline TurboQuant projection operates on the unpadded dimension $d=1536$.

\subsection{Execution Environment}
The projection operations for both methods execute as sequential loops, bypassing hardware vectorization libraries. A \ac{JIT} compiler translated the loops into single-threaded machine code. The setup removes low-level parallelization advantages, isolating the execution time of the $\mathcal{O}(d^2)$ matrix multiplication against the $\mathcal{O}(N \log_2 N)$ \ac{FWHT} butterfly network.

\subsection{Measurements}
The experiment tracks three metrics: the time to compute the intermediate continuous vector $\mathbf{y}$ for all database and query samples, the inner product distortion, and the retrieval accuracy. Following the projection, the pipeline quantizes the outputs to 1-bit representations using the sign function. The system scales the estimated inner products by $\frac{1}{d}\sqrt{\frac{\pi}{2}}$ for TurboQuant and $\frac{1}{N}\sqrt{\frac{\pi}{2}}$ for Fast-TurboQuant. We measured the \ac{MSE} between the 1-bit estimates and the unquantized ground truth $\mathbf{X}\mathbf{Q}^T$. We computed the Recall@10 metric to evaluate the intersection between the top-10 nearest neighbors retrieved using the 1-bit estimates against the unquantized baseline.

\subsection{Results and Analysis}
Table \ref{tab:results} summarizes the performance of the TurboQuant ($d=1536$) and the proposed Fast-TurboQuant ($N=2048$) under sequential execution.

\begin{table}
\centering
\caption{Benchmark on DBpedia OpenAI-3 Embeddings}
\label{tab:results}
\setlength{\tabcolsep}{3pt} 
\begin{tabular}{lcc}
\toprule
\textbf{Metric} & \textbf{TurboQuant ($d\!=\!1536$)} & \textbf{Fast-TurboQuant ($N\!=\!2048$)} \\
\midrule
Time (ms) & $2899.54$ & $147.08$ \\
MSE & $1.025 \times 10^{-3}$ & $9.09 \times 10^{-4}$ \\
Recall@$10$ & $0.7122$ & $0.7390$ \\
\bottomrule
\end{tabular}
\end{table}

Under unvectorized execution, Fast-TurboQuant completed in $147.08$ ms compared to $2899.54$ ms for TurboQuant. The $19.71$ times speedup aligns with the reduction in arithmetic operations.The Fast-TurboQuant architecture yielded an \ac{MSE} of $9.09 \times 10^{-4}$, operating below the TurboQuant \ac{MSE} of $1.025 \times 10^{-3}$. The error reduction is a byproduct of the \ac{FWHT} dimension constraints. From a rate-distortion perspective, zero-padding the native embedding from $d=1536$ to $N=2048$ expands the sketch dimension, increasing the bit budget per original coordinate. The dimension expansion provides the subsequent quantization stage with a larger capacity to encode the vector geometry, resulting in lower expected error.

We evaluate the relative ranking of the vectors using the Recall@10 metric. TurboQuant achieved a recall of $0.7122$, while the proposed projection achieved $0.7390$. The Rademacher-\ac{FWHT} pipeline preserves the semantic inner-product distances of the \ac{LLM} embeddings. The architecture satisfies the operational requirements for \ac{RAG} systems while eliminating the hardware footprint of the dense matrix.

\begin{figure}
    \centering
    \includegraphics[width=0.7\linewidth]{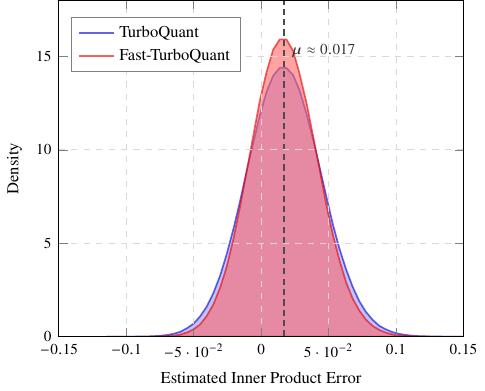}
    \caption{Empirical distribution of inner product estimation errors on DBpedia OpenAI-3 embeddings. The padded FJLT projection ($N=2048$) narrows the error distribution relative to the baseline projection ($d=1536$).}
    \label{fig:distortion_histogram}
\end{figure}

Figure \ref{fig:distortion_histogram} illustrates the density of the inner product estimation errors. Both projections yield distributions centered near zero, with a mean of $\mu\approx0.017$. However, the Fast-TurboQuant distribution exhibits a higher peak reaching $15.91$ compared to the baseline peak of $14.39$, featuring narrower tails. The distribution confirms that the multiplier-free pipeline preserves the statistical integrity of the inner products and improves the precision of the 1-bit quantization.

\acresetall

\section{Conclusion}

In this letter, we introduced Fast-TurboQuant, a multiplier-free projection architecture for online vector quantization. By replacing the dense Gaussian random rotation of the original TurboQuant architecture with a structured \ac{FJLT}, we demonstrated that the sub-Gaussian concentration of the Walsh-Hadamard butterfly network satisfies the prerequisites for 1-bit scalar quantization. 

This architectural substitution reduces the projection arithmetic complexity from $\mathcal{O}(d^2)$ to $\mathcal{O}(N \log_2 N)$. By leveraging a Rademacher diagonal matrix for phase inversion, the method eliminates floating-point multipliers from the projection data path, relying solely on sign-bit toggles and additions. 

Empirical evaluations on OpenAI-3 Large embeddings validate the theoretical framework. Under sequential execution, Fast-TurboQuant achieved a $19.71$ times algorithmic speedup over the native dense projection. Furthermore, the dimension expansion inherent to the power-of-two padding requirement of the \ac{FWHT} marginally reduces the error distribution, yielding an improved \ac{MSE} of $9.09 \times 10^{-4}$ and increasing the Recall@10 to $0.7390$. 

Ultimately, Fast-TurboQuant resolves the computational bottleneck associated with high-dimensional embedding compression. By matching the geometric preservation of dense random projections without the associated hardware penalty, the proposed architecture makes advanced \ac{RAG} systems and long-context \ac{KV} caches practical for deployment on constrained edge silicon.

\bibliographystyle{IEEEtran}
\bibliography{references}

\section*{Appendix A: Sub-Gaussian Concentration of the FJLT Projection}
\label{appendixproof}
Let $\mathbf{x} \in \mathbb{R}^d$ be a unit vector such as $\|\mathbf{x}\|_2 = 1$. The $i$-th coordinate of the projected vector $\mathbf{y} = \frac{1}{\sqrt{d}} \mathbf{H} \mathbf{D} \mathbf{x}$ is given by
\(
y_i = \frac{1}{\sqrt{d}} \sum_{j=1}^{d} W_j,
\)
where $W_j = H_{i,j} D_{j,j} x_j$.
Because $\mathbf{H} \in \{-1, 1\}^{d \times d}$ and $\mathbf{D}$ is a diagonal matrix of i.i.d. Rademacher variables, $W_j$ are independent, symmetric random variables taking values in $[a_j, b_j] = [-|x_j|, |x_j|]$. Consequently, $\mathbb{E}[W_j] = 0$.

By Hoeffding's general inequality for independent bounded variables, the unscaled sum $S_i = \sum_{j=1}^d W_j$ satisfies
\begin{equation}
\mathbb{P}(|S_i| \ge t) \le 2 \exp \left( - \frac{2t^2}{\sum_{j=1}^d (b_j - a_j)^2} \right).
\end{equation}
We evaluate the sum of the squared intervals
as \(\sum_{j=1}^d (b_j - a_j)^2 = 4\).
Substituting this into the inequality gives
\(
\mathbb{P}(|S_i| \ge t) \le 2 \exp \left( - \frac{t^2}{2} \right).
\)
Finally, substituting the scaled coordinate $y_i = S_i / \sqrt{d}$ and mapping the threshold $t = u\sqrt{d}$ yields the sub-Gaussian tail bound
\(
\mathbb{P}(|y_i| \ge u) \le 2 \exp\left(-\frac{d u^2}{2}\right).
\)

\end{document}